\pdfoutput=1
\documentclass[11pt]{article}
\usepackage[preprint]{acl} 

\usepackage{times}
\usepackage{latexsym}
\usepackage[T1]{fontenc}
\usepackage[utf8]{inputenc}
\usepackage{microtype}
\usepackage{inconsolata}
\usepackage{graphicx}

\usepackage{booktabs,multirow,multicol}
\usepackage{amsmath,amsfonts,bm}
\usepackage{xstring}

\usepackage{enumitem}

\usepackage{caption}
\usepackage{subcaption}

\usepackage{xcolor}
\definecolor{promptColorNone}{HTML}{1f77b4}
\definecolor{promptColorRandom}{HTML}{ff7f0e}
\definecolor{promptColorToxicity}{HTML}{2ca02c}
\definecolor{promptColorSentiment}{HTML}{d62728}
\definecolor{promptColorTopic}{HTML}{9467bd}
\definecolor{promptColorNLI}{HTML}{8c564b}

\usepackage{rotating}
\usepackage{makecell}
\newcommand{\vcell}[1]{\rotcell{\makebox[0pt][r]{#1}}}

\newcommand{\fref}[1]{Figure~\ref{#1}}
\newcommand{\tref}[1]{Table~\ref{#1}}

\newcommand{\cref}[1]{Chapter~\ref{#1}}
\newcommand{\sref}[1]{Section~\ref{#1}}
\newcommand{\aref}[1]{Appendix~\ref{#1}}

\newcommand{\std}[1]{\textsubscript{#1}}
\newcommand{\ignore}[1]{}

\DeclareMathOperator{\emb}{emb}

\newcommand{\significant}[1]{%
  \IfBeginWith{#1}{**}{**}{%
  \IfBeginWith{#1}{*}{*}{%
  \IfBeginWith{#1}{!!}{**}{%
  \IfBeginWith{#1}{!}{*}{%
  }}}}%
}
\newcommand{\pValNR}[2]{%
  \begin{tabular}[c]{@{\hspace{2pt}}l@{}}%
    \vspace{-3pt}\textcolor{blue}{\small{\significant{#2}}} \\
    \textcolor{violet}{\small{\significant{#1}}}
  \end{tabular}%
}
\newcommand{\best}[1]{\textbf{#1}}
\newcommand{\secondBest}[1]{\textit{#1}}

\title{Do Prompts Reshape Representations? \\ An Empirical Study of Prompting Effects on Embeddings}

\author{Cesar Gonzalez-Gutierrez \\
  Polytechnic University of Catalonia \\
  Barcelona, Spain \\
  \texttt{cesar.gonzalez.gutierrez@upc.edu} \\\And
  Dirk Hovy \\
  Bocconi University \\
  Milan, Italy \\
  \texttt{dirk.hovy@unibocconi.it} \\
}

\begin{document}

\maketitle

\begin{abstract}
Prompting is a common approach for leveraging LMs in zero-shot settings.
However, the underlying mechanisms that enable LMs to perform diverse tasks without task-specific supervision remain poorly understood.
Studying the relationship between prompting and the quality of internal representations can shed light on how pre-trained embeddings may support in-context task solving.
In this empirical study, we conduct a series of probing experiments on prompt embeddings, analyzing various combinations of prompt templates for zero-shot classification.
Our findings show that while prompting affects the quality of representations, these changes do not consistently correlate with the relevance of the prompts to the target task. This result challenges the assumption that more relevant prompts necessarily lead to better representations.
We further analyze potential factors that may contribute to this unexpected behavior.
\end{abstract}

\section{Introduction}
\label{intro}

In recent years, language model (LM) prompting has emerged as the dominant model application paradigm in NLP, with LMs like GPT-3 \citep{brown_language_2020}, LLaMA \citep{touvron2023llama}, PaLM \citep{chowdhery-etal-2024-palm}, among many others.
In this framework, the model architecture remains unchanged, and the downstream task is verbalized transforming tasks into text prompts which are inputted to the model to elicit the desired response. Prompting leverages LMs performing language modeling tasks through conditional text generation or clause-style completion.

A key advantage of this approach is its generality. A single model architecture can ideally handle various tasks without further supervised training, eliminating the need for task-specific fine-tuning. This method also offers great flexibility, allowing tasks to be switched simply by changing the prompt. This could in principle enable the model to address tasks it was not explicitly trained on, a phenomenon often referred to as model's \textit{emergent abilities} \citep{brown_language_2020, schaeffer-etal-2023-emergent}.
Although LMs have high computational demands, prompting also offers advantages in low-annotation scenarios, by leveraging their encoded knowledge to address under-resourced tasks \citep{mosbach-etal-2023-shot}.

The model's ability to solve a task processing a prompt containing the target sample is known as in-context learning (ICL).
Tasks can be approached including solved examples in the prompt (\textit{few-shot} learning), or by directly instructing the model without explicit examples (\textit{zero-shot} learning). Contrary to supervised training, ICL does not need to update the model weights, leveraging LM's pre-training without further adaptation, and achieving performance solely by contextualizing the sample in a prompt.

For a prompt to be effective for a task, the language used in the context of such task must have been encountered during pre-training \citep{gonen2024demystifying}. This exposure would allow the model to learn patterns and recognize language structures relevant to the particular task.
Conversely, if a LM has not seen patters analogous to those relevant for a given task, this should result in lower performance when processing the prompt.

As an example, similar to the argument made by \citet{radford2019language} on how ICL might work in practice, consider the prompt ``Is the following review positive or negative?: The movie was great!''. In this case, the review itself is the sample to evaluate. The question, explicitly states the task objective (sentiment analysis) and refers to the review for evaluation.
The language patterns that appear while expressing or commenting on others opinions (e.g., ``Did you enjoy the movie? I think it was awesome.''), although formulated less explicitly, have a similar structure. These patters are common in language and their structure can be learned by LMs during pre-training. This illustrates how a LM can build useful representations from seemingly unrelated contexts for ICL, provided that the corpus is sufficiently large and general.

If LMs are capable of performing a task without modifying the pre-trained weights, then all that occurs during ICL are representational changes via processing of the prompt tokens.
At the representation level, ICL can be understood as contextualization of samples via prompting. This raises the question: What happens to the sample representation when we contextualize it in a prompt?
Can we measure any improvement in the representation when it's contextualized in a prompt that makes it more suitable for a particular task?

If ICL is driven by language patterns shared between a task and the data seen during pre-training, then prompts that are relevant to the task should improve the quality of the resulting representations. In contrast, irrelevant prompts should produce no improvement or even worsen quality. Based on this assumption, we hypothesize that changes in representational quality induced by prompting (relative to a specific task) will align with the relevance of the prompt to the task. Moreover, such representational changes should be measurable using representation analysis tools.

Additionally, if ICL capabilities in transformer models result solely from language modeling tasks, then these capabilities should not depend on the specific type of pre-training (as long as the model can effectively model the input distribution). In our experiments, we test quality changes using both masked language models (MLM) and autoregressive models. This property should also be present in contextual models of any size, and its effects detectable in smaller models.
Therefore, if the effects are significant, we should be able to study them in models with only a few hundred million parameters, small enough to run on mid-sized hardware.

In this empirical study, we employ probing techniques \citep{ettinger-etal-2016-probing, adi2017finegrained} to analyze the task-relevant information encoded within sentence representations. Specifically, we compute task-specific probe performance on prompt embeddings, computed from the target samples along task-specific instructions (zero-shot prompting). We then compare probe performance between different prompt templates, some being relevant to the probe task while others unrelated.

In summary, this work studies the effect of prompting at the embedding level. We hypothesize that prompting contextualizes sentence representations, leading to measurable changes in their embeddings, as reflected in probing task performance. We expect probe performance to improve when prompts are relevant to the task, and decrease when they are irrelevant.
We anticipate that similar effects may also be observed using alternative metrics of task-specific representation quality.

The main contributions of this paper are:
\begin{itemize}[topsep=0em, parsep=0em, itemsep=0em, leftmargin=1em]
    \item We conduct an empirical analysis comparing the quality of sentence representations across several tasks, using prompt templates that are either relevant or irrelevant to the target task.
    \item We show that prompting alters sample's sentence representations through contextualization.
    \item We find that changes in prompt embedding quality due to prompt relevance do not follow a consistent or predictable pattern across tasks and models.
\end{itemize}

\section{Experimental Setup}

\subsection{Learning Setting}
\label{learning-setting}

\paragraph{Prompting}
In a prompting pipeline \citep{liu-etal-2023-pretrain}, the input text $x$ is first modified using a prompting function $p$. This function applies a template or prefix to produce a prompt $x' = p(x)$. The prompt may include several examples of the task (few-shot learning), or none (zero-shot learning), along with task instructions.

This prompt is then used as input to the model for language modeling, such as masked token prediction or next token prediction.
The highest-scoring answer is selected from the model output, according to some LM criteria, for example, by maximizing MLM token probability.
Finally, the selected answer is mapped to the best annotation from the set of possible labels $\hat{y} \in \mathcal{Y}$ through some task-specific criteria, e.g., highest similarity to a set of predefined answers.

\paragraph{Representations}
We are interested in the effect of prompting on sentence-level representations in a zero-shot learning setting. To study this scenario, we will generate text representations using an embedding function $\emb: \mathcal{X} \rightarrow \mathbb{R}^n$, that maps a text fragment to a vector space. There are several strategies to build embedding representations based on the activations of a model $\mathcal{M}(\cdot; \theta)$ applied to the input, using the generated token representations at different layers \citep{reimers-gurevych-2019-sentence, devlin-etal-2019-bert}. To measure the effect prompting has on representations, we will consider various task-specific prompts $p_\text{task}$, apply them for each dataset sample $x' = p_\text{task}(x)$, and generate their embedding representations: $\bm{r} = \emb(x')$.

\paragraph{Prompting Effect Analysis}
If prompting benefits a task and this is reflected at the representation level, then probing should reveal performance differences between prompts, whether they are relevant or irrelevant to the task. To test this, we probe the representations of task-specific prompts applied to various datasets and compare the resulting performance. We use the unmodified input as a primary baseline. To control for potential spurious effects caused by simply adding tokens, as second baseline we include a template with a random list of words.

\subsection{Measuring Prompt Embedding Quality}
\label{method}

\paragraph{Classification Tasks}
In this study, we examined four classification tasks. For \textit{toxicity detection}, we used Wiki Toxic \citep{wulczyn-etal-2017-ex}. For \textit{sentiment analysis}, we considered IMDB \citep{maas-etal-2011-learning}. For \textit{topic classification}, we used AG News \citep{zhang2015character}, News Articles\footnote{\href{https://huggingface.co/datasets/valurank/News_Articles_Categorization}{huggingface.co/datasets/valurank/ News\_Articles\_Categorization}}, Arise News\footnote{\href{https://huggingface.co/datasets/okite97/news-data}{huggingface.co/datasets/okite97/news-data}}, and Swahili News \citep{david_2020_swahili}. For natural language inference (\textit{NLI}), we used GLUE's RTE benchmark \citep{wang-etal-2018-glue}\footnote{\href{https://aclweb.org/aclwiki/Recognizing_Textual_Entailment}{aclweb.org/aclwiki/Recognizing\_Textual\_Entailment}} and the second round of the Adversarial NLI dataset \citep{nie-etal-2020-adversarial}.
Further datasets details can be found in \aref{datasets}.

\paragraph{Prompt Templates}
We focused on prompts designed for zero-shot ICL.
We designed five prompt templates per task, plus five random prompts and the unmodified input, totaling 26 templates (detailed in \tref{tab:prompts}).
Each template is applied to the dataset samples by substitution before feeding it to the model.
Whenever a template required one input but the dataset had two (e.g., a sentiment prompt with an NLI example), we simply concatenated the inputs. Conversely, for NLI prompts with only one input text, we used the premise position alone.

\begin{table}[tb]
\centering
\begin{small}
\begin{tabular}{@{\hspace{0em}}c@{\hspace{0.2em}}r@{\hspace{0.8em}} p{20.5em}@{\hspace{0em}}}
\toprule
Task & \# & Prompt \\
\midrule
\multicolumn{2}{l}{None} & \textit{\{text\}}\\
\midrule
\multirow[c]{4}{*}{\vcell{Toxicity}} & 1 & Is this a toxic comment?: \textit{\{text\}} \\
 & 2 & Should we avoid this type of comment in a respectful conversation?: \textit{\{text\}} \\
 & 3 & Do you consider this comment to be respectful?: \textit{\{text\}} \\
 & 4 & As a moderator, would you remove such a comment?: \textit{\{text\}} \\
 & 5 & Check for toxicity: \textit{\{text\}}\\
\midrule
\multirow[c]{3}{*}{\vcell{Sentiment}} & 1 & Is the sentiment of this comment positive or negative?: \textit{\{text\}} \\
 & 2 & Analyze this comment to determine the overall sentiment: \textit{\{text\}} \\
 & 3 & Evaluate the overall sentiment in this comment: \textit{\{text\}} \\
 & 4 & What emotions are conveyed in this text?: \textit{\{text\}} \\
 & 5 & Sentiment Analysis: \textit{\{text\}}\\
\midrule
\multirow[c]{6}{*}{\vcell{Topic}} & 1 & Classify the following text based on its main  topic: \textit{\{text\}} \\
 & 2 & Is this text about Sports, World news, Business or Science?: \textit{\{text\}} \\
 & 3 & Determine whether the text belongs to Finance or Sports: \textit{\{text\}} \\
 & 4 & Which category do you believe best summarizes the main topic of this text? \textit{\{text\}} \\
 & 5 & Topic classification: \textit{\{text\}}\\
\midrule
\multirow[c]{3}{*}{\vcell{NLI}} & 1 & Given that: \textit{\{premise\}}. It is true that: \textit{\{hypothesis\}} \\
 & 2 & Given the sentence \textit{\{premise\}}, determine if the following statement is entailed: \textit{\{hypothesis\}} \\
 & 3 & If \textit{\{premise\}}, then \textit{\{hypothesis\}} \\
 & 4 & It is the case that \textit{\{hypothesis\}}, because \textit{\{premise\}} \\
 & 5 & \textit{\{premise\}}, which means that \textit{\{hypothesis\}}\\
\midrule
\multirow[c]{5}{*}{\vcell{Random}} & 1 & Spiky hospital aspiring tooth scale?: \textit{\{text\}} \\
 & 2 & Abandoned questionable converts silent available cup dance belligerent \textit{\{text\}} \\
 & 3 & knowing resolve profit giddy spiteful songs guide attractive fancy large \textit{\{text\}} \\
 & 4 & bustling like innate face important grind stretch rhythm: \textit{\{text\}} \\
 & 5 & detailed operate channel sweet hands uninterested turn addition: \textit{\{text\}} \\
\bottomrule
\end{tabular}
\end{small}
\caption{Prompt templates for different tasks.}
\label{tab:prompts}
\end{table}

\begin{figure*}[htbp]
\centering
\begin{subfigure}{\linewidth}
    \includegraphics[width=\linewidth]{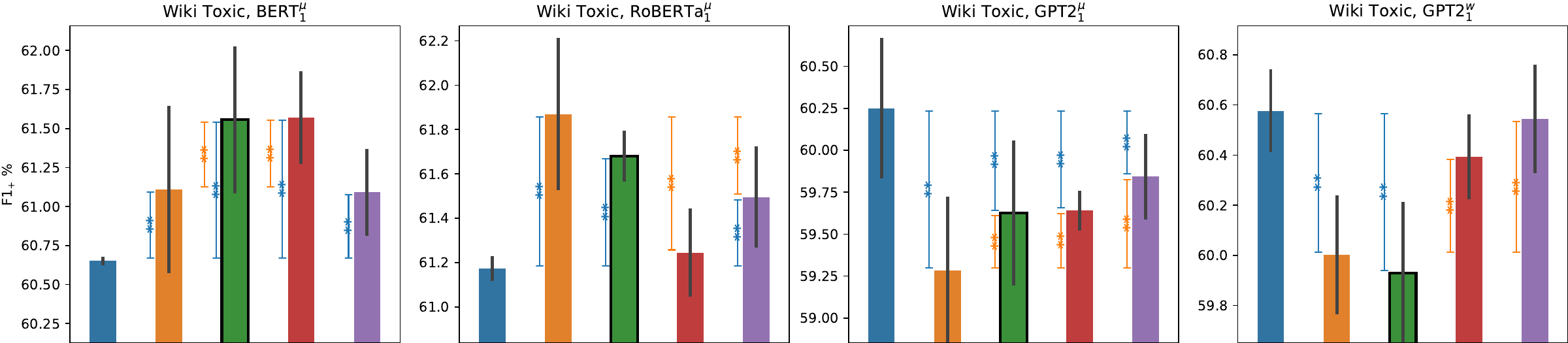}
    \caption{{\bf\color{promptColorToxicity}Toxicity} detection (Wiki Toxic) with varying prompt representations.}
    \label{fig:probing-wt}
    \vspace{0.5em}
\end{subfigure}
\begin{subfigure}{\linewidth}
    \includegraphics[width=\linewidth]{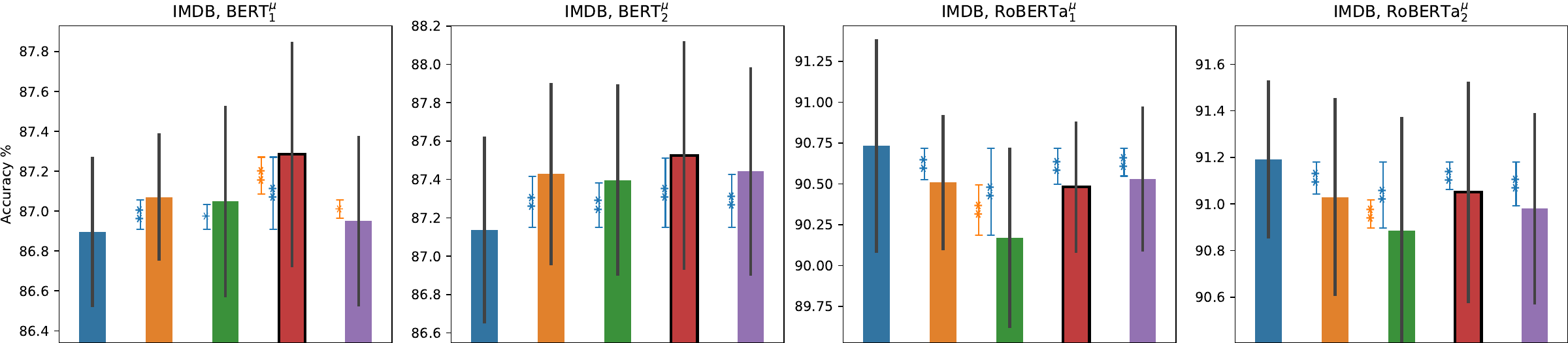}
    \caption{{\bf\color{promptColorSentiment}Sentiment} classification (IMDB) with varying prompt representations.}
    \label{fig:probing-imdb}
    \vspace{0.5em}
\end{subfigure}
\begin{subfigure}{\linewidth}
    \includegraphics[width=0.49\linewidth]{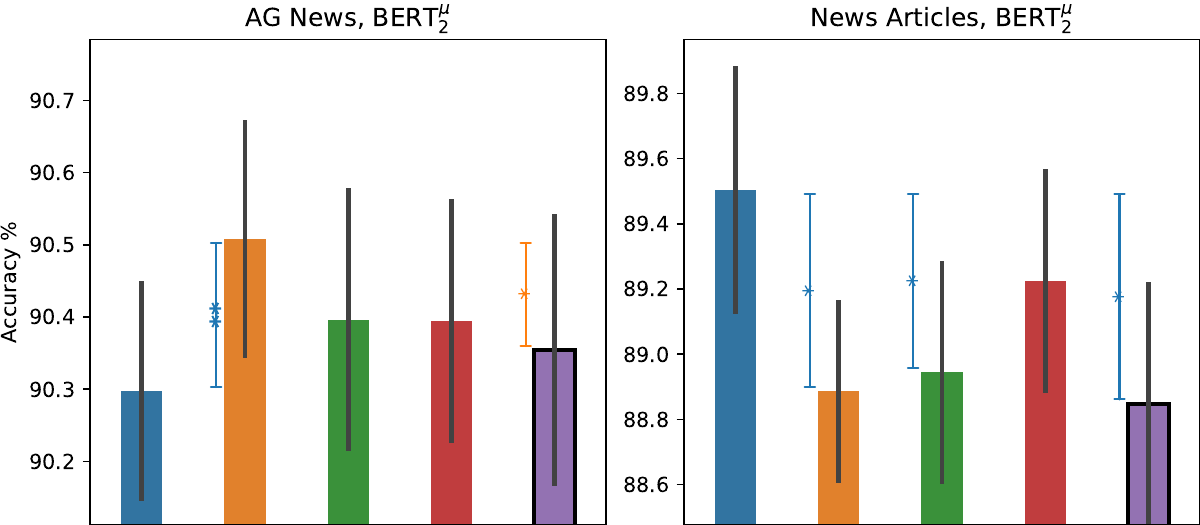}\hfill%
    \includegraphics[width=0.49\linewidth]{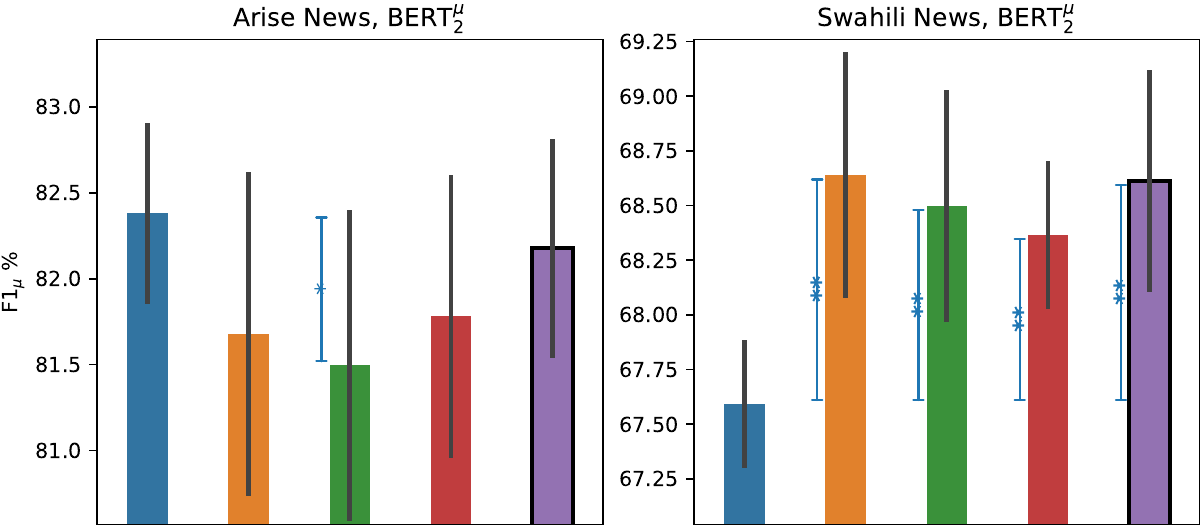}
    \caption{{\bf\color{promptColorTopic}Topic} classification datasets using a fixed prompt representation.}
    \label{fig:probing-news}
    \vspace{0.5em}
\end{subfigure}
\begin{subfigure}{\linewidth}
    \includegraphics[width=\linewidth]{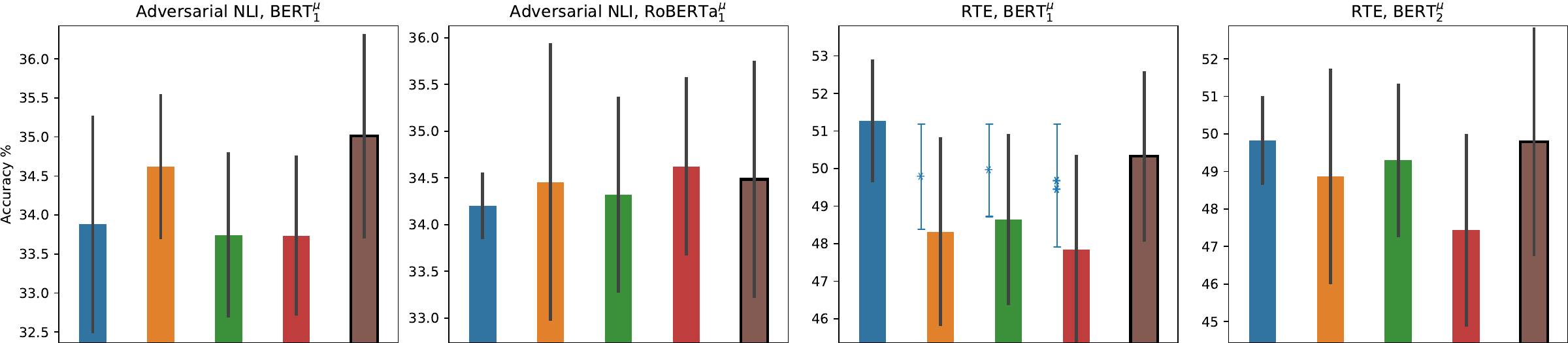}
    \caption{Combinations of {\bf\color{promptColorNLI}NLI} datasets and prompt representations.}
    \label{fig:probing-topic}
    \vspace{0.5em}
\end{subfigure}
\includegraphics[width=\linewidth]{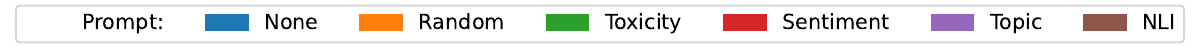}
\caption{Probe performance of prompt embeddings. The task-relevant prompt performance is indicated using black borders.
We report statistical significance lines w.r.t. no prompt (blue) and random prompt (orange). We consider two significance levels: $p < 0.05$ and $p < 0.01$.}
\label{fig:probing}
\end{figure*}

\paragraph{Models}
For representation generation, we selected three models with varying pre-training length and language modeling tasks. We used BERT \citep{devlin-etal-2019-bert}, which combines MLM with next-sentence prediction; RoBERTa \citep{liu2019roberta}, a model trained exclusively on MLM over a larger corpus; and GPT-2 \citep{radford2019language}, a left-to-right autoregressive model trained on a general corpus.

\paragraph{Embeddings}
We then generated embedding representations by applying all prompt templates to the dataset samples, resulting in five prompt embeddings per task, along with the embedding of the original input.
When generating these embeddings, we considered various pooling strategies at both layer and token levels. For BERT and RoBERTa MLM models, we used either the last ($_1$) or the second-to-last ($_2$) layer \citep{devlin-etal-2019-bert}. For token pooling, we either averaged all token representations ($^\mu$) or used the \texttt{[CLS]} token ($^\texttt{CLS}$). For GPT-2, we used the last layer and either an average token representation or a weighted average ($^w$), with left-to-right linearly increasing weights.

\paragraph{Probes}
To probe the representations for encoded task information, we trained MaxEnt classifiers with L2 regularization on top of the representation and tested their performance on the test partition.

\paragraph{Statistics}
Differences in probe performance among the prompt representations might be subtle, making the benefits of prompting less apparent. Therefore, we studied the statistical significance of those differences with bootstrap sampling statistics \citep{berg-kirkpatrick-etal-2012-empirical, sogaard-etal-2014-whats}, using the \textit{boostsa} library \citep{fornaciari-etal-2022-hard}. We computed p-values of probe performance relative to both the original example and the random prompts.

\section{Results}
\label{results}

\subsection{Probing Performance of Prompt Embeddings}
\label{results_main}

\fref{fig:probing} presents the main probing results obtained using the templates from \tref{tab:prompts} grouped by target task. These were tested on probing tasks corresponding to the target tasks of the prompts, using different datasets, model architectures and embedding strategies.

The model architecture used to generate representations has a notable impact on probe performance across prompts. For the Wiki Toxic and IMDB datasets, using the original input as a baseline, BERT generally shows statistically significant improvements with any prompt (including random ones). In contrast, RoBERTa's behavior varies by dataset, and GPT-2 consistently shows degraded performance. Compared to the random prompt baseline, probe performance changes with model architecture, where only BERT demonstrates significant gains when using the relevant prompt.

However, for topic classification tasks, BERT representations fail to achieve significant improvements even with the relevant prompt. Each dataset exhibits a distinct performance pattern, suggesting these differences are not tied to the task type or representation method alone.

In the case of NLI tasks, prompt representation quality tends not to significantly impact probe performance when compared to the unchanged input or using a random prompt template. An exception is observed with the RTE dataset using BERT's final layer: here, performance declines with most prompts except the relevant one.

Overall, probe performance of prompt representations is highly dependent on the model and dataset.
There is no consistent performance pattern across tasks and prompts that supports our initial hypothesis: that a sufficiently pre-trained model would benefit from relevant task prompts, reflecting this improvement at the representation level. The empirical evidence does not support this hypothesis.
In some cases, prompts for unrelated tasks can improve probe performance, while, in other cases, relevant prompts may even degrade performance. Additionally, random prompts can also enhance performance, contrary to intuition, echoing findings from prior work \citep{lu-etal-2024-strings}.

\subsection{Ablation Studies}
\label{ablation}

In this section, we extend the previous study to better understand the factors influencing the behavior observed in prompt representations.

\begin{table*}[hbt]
\centering
\begin{tabular}{ll lllll}
\toprule
\multirow{2.5}{*}{Dataset} & \multirow{2.5}{*}{Repr.} & \multicolumn{5}{c}{Prompt} \\
\cmidrule(r){3-7}
 & & None & Random & Toxicity & Sentiment & Topic \\
\cmidrule(lr){1-7}

\multirow{6}{*}{Wiki Toxic} & BERT$_1^{\mu}$ & \best{96.82}\std{0.12} & 96.45\std{0.14} & \secondBest{96.70}\std{0.12} & \secondBest{96.70}\std{0.10} & 96.63\std{0.14} \\
 & BERT$_2^{\mu}$ & \best{96.65}\std{0.08} & 96.30\std{0.17} & 96.59\std{0.12} & \secondBest{96.62}\std{0.10} & 96.52\std{0.15} \\
 & RoBERTa$_1^{\mu}$ & \best{96.79}\std{0.15} & 96.56\std{0.12} & \secondBest{96.58}\std{0.10} & 96.56\std{0.12} & 96.47\std{0.14} \\
 & RoBERTa$_2^{\mu}$ & \best{96.89}\std{0.09} & \secondBest{96.68}\std{0.13} & \secondBest{96.68}\std{0.10} & 96.65\std{0.12} & 96.59\std{0.13} \\
 & GPT-2$_1^{\mu}$ & \best{95.44}\std{0.17} & 94.78\std{0.20} & 95.01\std{0.15} & \secondBest{95.02}\std{0.16} & 94.99\std{0.19} \\
 & GPT-2$_1^w$ & \best{95.49}\std{0.13} & 94.96\std{0.22} & \secondBest{95.24}\std{0.14} & \secondBest{95.24}\std{0.18} & 95.20\std{0.19} \\
\cmidrule(lr){1-7}

\multirow{4}{*}{IMDB} & BERT$_1^{\mu}$ & 88.42\std{0.13} & 88.48\std{0.13} & 88.42\std{0.12} & \best{88.56}\std{0.11} & \secondBest{88.55}\std{0.11} \\
 & BERT$_2^{\mu}$ & 88.36\std{0.12} & 88.66\std{0.13} & 88.63\std{0.14} & \best{88.72}\std{0.11} & \secondBest{88.71}\std{0.13} \\
 & RoBERTa$_1^{\mu}$ & 87.73\std{0.20} & 87.90\std{0.25} & 87.83\std{0.19} & \best{87.94}\std{0.19} & \secondBest{87.92}\std{0.20} \\
 & RoBERTa$_2^{\mu}$ & 88.76\std{0.23} & \secondBest{88.92}\std{0.21} & 88.85\std{0.21} & \best{88.94}\std{0.24} & 88.89\std{0.15} \\
\cmidrule(lr){1-7}

\multirow{4}{*}{AG News} & BERT$_1^{\mu}$ & \best{94.87}\std{0.13} & 94.81\std{0.15} & 94.82\std{0.14} & \secondBest{94.85}\std{0.12} & 94.78\std{0.13} \\
 & BERT$_2^{\mu}$ & 94.58\std{0.15} & \secondBest{94.61}\std{0.16} & \secondBest{94.61}\std{0.17} & \best{94.67}\std{0.13} & \secondBest{94.61}\std{0.14} \\
 & RoBERTa$_1^{\mu}$ & \secondBest{94.00}\std{0.18} & \best{94.13}\std{0.23} & \secondBest{94.00}\std{0.17} & 93.96\std{0.19} & 93.89\std{0.22} \\
\vspace{0.5em} & RoBERTa$_2^{\mu}$ & \secondBest{94.20}\std{0.20} & \best{94.28}\std{0.23} & 94.15\std{0.23} & 94.16\std{0.19} & 94.10\std{0.21} \\
News Art. & BERT$_2^{\mu}$ & 93.57\std{0.00} & 93.69\std{0.08} & 93.68\std{0.11} & \secondBest{93.70}\std{0.05} & \best{93.71}\std{0.07} \\
Arise News & BERT$_2^{\mu}$ & 92.59\std{0.00} & 92.72\std{0.06} & \secondBest{92.89}\std{0.10} & \best{92.90}\std{0.09} & 92.73\std{0.12} \\
Swh. News & BERT$_2^{\mu}$ & \best{85.19}\std{0.07} & 85.04\std{0.12} & 85.17\std{0.14} & \best{85.19}\std{0.16} & \secondBest{85.18}\std{0.20} \\

\cmidrule(lr){1-7}
 & & None & Random & Toxicity & Sentiment & NLI \\
\cmidrule(r){3-7}
\multirow{4}{*}{Adv. NLI} & BERT$_1^{\mu}$ & 81.54\std{0.22} & \best{81.57}\std{0.22} & \secondBest{81.56}\std{0.22} & \secondBest{81.56}\std{0.22} & \secondBest{81.56}\std{0.22} \\
 & BERT$_2^{\mu}$ & 81.54\std{0.21} & \best{81.58}\std{0.21} & \best{81.58}\std{0.22} & \secondBest{81.57}\std{0.22} & 81.56\std{0.22} \\
 & RoBERTa$_1^{\mu}$ & 81.69\std{0.24} & \secondBest{81.73}\std{0.24} & 81.72\std{0.24} & 81.72\std{0.24} & \best{81.81}\std{0.26} \\
 & RoBERTa$_2^{\mu}$ \vspace{0.5em} & 81.78\std{0.24} & \secondBest{81.83}\std{0.26} & \secondBest{81.83}\std{0.26} & \secondBest{81.83}\std{0.26} & \best{81.94}\std{0.27} \\
 RTE & BERT$_2^{\mu}$ & 79.59\std{0.00} & \best{79.86}\std{0.10} & \best{79.86}\std{0.09} & \secondBest{79.76}\std{0.03} & 79.66\std{0.19} \\
\bottomrule
\end{tabular}
\caption{Task Alignment of prompt representations, with standard deviation reported in subscripts.}
\label{tab:alignment}
\end{table*}

\subsubsection{Representation Choice}
\label{representations}

In this extended experimental study, we consider a broader range of representation generation strategies to understand how they influence probe performance. We include additional token representations and pooling mechanisms within the previously discussed architectures.
Specifically, we consider the \texttt{[CLS]} token to evaluate its importance in prompt-based embeddings and analyze the behavior of single token representations. 

The results presented in \tref{tab:probing} (in \aref{extra}) show that the choice of representation strategy introduces another source of variability in probe performance. Different strategies can lead to varying results, even within the same architecture.
Additionally, representations based on the \texttt{[CLS]} token exhibit different behavior compared to those using average pooling, and they generally perform worse than their averaged counterparts.

\subsubsection{Task Alignment}
\label{ta}

To complement the results in \sref{results}, this set of experiments uses an alternative metric to evaluate representation quality. Specifically, we study whether prompts induce parallel changes in task alignment \citep{gonzalez-gutierrez-etal-2023-analyzing}, as observed with probing. Task alignment measures the degree of agreement between the representation space and the task space. This score is calculated as the average across all levels of clustering granularity of partition classification scores obtained by label probabilities proportional to in-cluster class prevalence. This score was computed including all the combinations of prompts and tasks reported in \tref{tab:probing} (see \aref{extra}).

The results are shown in \tref{tab:alignment}. We find that prompts influence task alignment in a manner similar to their effect on probing, without any consistent or predictable pattern with respect to prompt relevance. The only exception is the sentiment classification dataset.

\tref{tab:alignment_correlation} shows the correlation coefficients between task alignment scores and probing performance. There is a strong positive correlation between the two metrics, suggesting that the observed behavior can be traced back to changes in the class distribution within the embedding space induced by prompt instructions. However, this alternative measure does not provide further insight into the unexpected behavior of relevant versus irrelevant prompts.

\begin{table}[htb]
\centering
\begin{tabular}{l rr}
Pearson's & $r = 0.7475$ & \small{$p < 10^{-19}$} \\
Spearman's & $\rho = 0.8412$ & \small{$p < 10^{-28}$} \\
Kendall's & $\tau = 0.6651$ & \small{$p < 10^{-23}$} \\
\end{tabular}
\caption{Task Alignment vs. Probe Performance correlation coefficients.}
\label{tab:alignment_correlation}
\end{table}

\begin{table*}[htb]
\centering
\begin{tabular}{c llllll}
\toprule
\multirow{2.5}{*}{Repr.} & \multicolumn{6}{c}{Prompt} \\
\cmidrule(lr){2-7}
 & None & Random & Toxicity & Sentiment & Topic & NLI \\
\cmidrule(lr){1-7}
BERT$_1^{\mu}$ & 60.47 & 60.48\pValNR{0.353}{} & 60.49\pValNR{0.1525}{0.2185} & 60.48\pValNR{0.3545}{0.446} & 60.47\pValNR{0.5445}{0.602} & 60.47\pValNR{0.5235}{0.5805} \\
BERT$_2^{\mu}$ & 60.11 & 60.13\pValNR{*0.01}{} & 60.11\pValNR{0.2245}{0.825} & 60.13\pValNR{*0.0315}{0.391} & 60.12\pValNR{0.0945}{0.6525} & 60.10\pValNR{0.566}{0.949} \\
\bottomrule
\end{tabular}
\caption{Probe scores of representations using static prompts for the Wiki Toxic dataset.}
\label{tab:static}
\end{table*}

\subsubsection{Prompt Structure}
\label{mask_sep}

A prompt consists of a template filled with the sample of interest and is comprised of two parts: the sample itself and the accompanying text, usually in the form of task instructions.
These two components operate at different levels of language: the object language, which contains the target sample, and the meta-language, which describes the task and refers to the sample. The representation of these two parts may relate differently to the task.
In this section, we analyze how each component of the prompt contributes to probe performance and compare their respective behaviors.

Intuitively, in an LM representation, the instruction tokens modify the sample tokens through contextualization, and vice versa. However, in an MLM architecture, once the context has been established, the instruction tokens are not necessarily required for forming the final sentence representation. The instruction tokens themselves might negatively impact the sentence representation if they introduce noise as being semantically distant with respect to the sample.
In our first set of experiments, we constructed sentence representations considering only the tokens corresponding to the sample (including \texttt{[CLS]}), while masking the remaining tokens in the template. The results are shown in \aref{extra}, \tref{tab:mask_sep} (top).

BERT also supports pairs of sentences separated by the special token \texttt{[SEP]}. During pre-training, these pairs are used for the next-sentence prediction objective. This design choice benefits tasks naturally represented as pairs of sentences (e.g., NLI), by enhancing the model’s ability to capture sentence relationships \citep{devlin-etal-2019-bert}. This raises the question: can we leverage this architecture to better distinguish between instructions and the sample, and thereby improve prompt representations overall?

To explore this, we conducted a second set of experiments where we used BERT’s \texttt{[SEP]} token to separate the instructions from the sample. The results are presented in \aref{extra}, \tref{tab:mask_sep} (middle). We also evaluated a combination of both techniques, using a masked prompt and separating instructions with \texttt{[SEP]}, shown at the bottom of the same table.

The probing results from these experiments do not reveal any qualitatively different behavior in representation quality. Whether using the masked prompt, the separator, or both, only slight changes in probe performance were observed. As with the main experimental study, there remains no consistent relationship across tasks or prompt relevance.
From the first set of experiments (\tref{tab:mask_sep}, top), we conclude that the contextualized sample tokens are sufficient to build a representation. Additionally, using sentence pairs separated by \texttt{[SEP]} does not improve the alignment between task and prompt relevance. We are therefore unable to leverage this architectural feature to better distinguish between the sample and the instructions.

\subsubsection{Static Prompt}
\label{static}

When using static token representations, we do not expect prompts to improve the quality of the resulting embeddings. Intuitively, only contextualized representations can benefit from prompts in a task-solving context, as instructions and samples can influence token embeddings in some useful way. As a base experiment, we test whether static prompts can influence probe performance. We assess whether a linear combination of the template and sample representations can improve their quality.

To construct the prompt representations, we first generated sentence embeddings for the samples. Separately, we computed an embedding of the template instructions. The final sentence embeddings were obtained by averaging the sample and instruction embeddings. This experiment was conducted using only the Wiki Toxic dataset.

The results in \tref{tab:static} show that this construction method effectively neutralizes the effect of prompting.
Probe performance across prompts disappear and become statistically insignificant.
At the embedding level, using a static prompt amounts to applying a spatial translation to the sample representation.
As expected, this has no meaningful impact on probe performance.
Therefore, prompt instructions must influence token representations trough contextualization within the model in order to be effective.

\section{Related Work}
\label{related_work}

\paragraph{Representation Analysis}
The literature on probing linguistic capabilities of representations is extensive \citep{belinkov-glass-2019-analysis}. \citet{hewitt-manning-2019-structural, reif_visualizing_2019} find syntactic structures latent in the vector space. \citet{miaschi-dellorletta-2020-contextual} studied dependencies, and how the ability to contextualize word embeddings can be applied to sentence embeddings. Other researchers have studied agreement \citep{hanna-etal-2023-functional}, grammaticality \citep{marvin-linzen-2018-targeted}, sentence structure \citep{tenney_what_2019}, or recursivity \citet{lyu-etal-2022-favorite}.
Representation learning dynamics has also been explored across various syntactic \citep{chiang-etal-2020-pretrained, saphra-lopez-2019-understanding}, semantic \citep{liu-etal-2021-probing-across, liu-etal-2019-linguistic, muller-eberstein-etal-2023-subspace}, or multilingual model capabilities \citep{wang-etal-2024-probing-emergence, blevins-etal-2022-analyzing}.

An active area of research within representation analysis is the study of transformer circuits. This research program aims to achieve \textit{mechanistic interpretability} by reverse-engineering transformers to identify functional units as interpretable computational structures. \citet{elhage2021mathematical} introduced the concept of induction heads, specialized attention heads which predict next tokens by induction from past sequences. Building on this idea, \citet{olsson2022context} analyzed the importance of induction heads in enabling in-context learning.

\paragraph{Understanding In-Context Learning}
Other works try to explain the model mechanisms that enable ICL taking an algorithmic perspective. \citet{garg2022what} studied the function classes that ICL is capable of learning. \citet{todd2024function} describe a mechanism of autoregressive models similar to function application. Other works have described this phenomenon as a meta-algorithm in the model activations, such as gradient descent \citep{akyurek2023learning, vonoswald2023transformers}.

Among the studies more closely related to our objectives is \citet{park2025iclr}, which studies the ability of LMs to produce new representations in-context. In contrast, our work focuses on how models can improve existing representations through prompting. \citet{kirsanov2025geometry} also studies representational changes induced by prompting, but their focus is on measuring class separability in large autoregressive models using synthetic datasets. Our approach, by contrast, seeks to uncover prompting mechanisms that are common across transformer architectures, using classic representation analysis tools.

\section{Conclusion and Future Work}
\label{conclusion}

By applying probes to the embedding representations of prompts, we observed that prompting modifies sentence-level representations not only by introducing new tokens but mainly by contextualizing the tokens of the original sample. We found that such changes amount to a redistribution of class samples in the embedding space.

However, our experimental study does not clearly explain the mechanisms that enable zero-shot ICL through prompting in LMs. Differences in probe performance between task-specific prompts shows no consistent patterns. In particular, we cannot conclude that a prompt relevant to a target task improves the representations generated by the model, as initially hypothesized. Additionally, as noted in previous research \citep{lu-etal-2024-strings}, seemingly irrelevant changes to the input text can lead to unexpected performance variations.

The available results do not clearly explain why this behavior occurs. One possibility is that the embedding-level perspective is too limited to capture the complexities of ICL, where the layer dynamics of input processing may play a crucial role. Another possibility is that the models used in our experiments were not sufficiently pre-trained to support effective prompting. The size of the pre-training corpora used in current state-of-the-art models is significantly larger than that used in our study.

Another possibility is that pre-training alone may not be enough for models to perform well with prompts.
LMs are often further adapted through supervised learning, such as instruction fine-tuning or reinforcement learning from human feedback, to improve their responsiveness to user queries. 
This additional training may be necessary to achieve stronger ICL performance. Our results do not allow us to determine whether better pre-training data or supervised adaptation would lead to representation improvements under prompting.

An enhanced experimental study would be necessary to better understand the behavior of representations under ICL and why embedding changes are not aligned with the prompt relevance in this particular learning setting.

\section*{Limitations}

This work focuses on analyzing representational changes caused by prompting that may be common across transformer architectures, serving as a baseline for identifying such behavior in these models. Our analysis of embedding spaces adopts a static view of the representations generated by language models. However, this perspective may not be sufficient to fully explain the phenomenon of in-context learning. To capture the more complex dynamics that occur during prompt processing, a different approach may be necessary: one that considers the model’s internal computations and the evolving nature of token interactions throughout the forward pass.

The LMs used in our experiments were pre-trained on relatively small corpora compared to those used for modern large-scale models. Although this is an experimental design choice aiming to find prompting effects in smaller models, this limited pre-training may not be sufficient for the models to fully develop the capabilities needed to benefit from prompting. As a result, our findings may not generalize to larger, instruction-tuned models that have been shown to exhibit more robust prompt-driven behavior.

Our analysis focused on a limited set of classification tasks and datasets, such as toxicity detection, sentiment analysis, and topic classification. The generalizability of our findings to other tasks, especially those that lie in more complex output spaces, remains an open question.

\anonymize{
\section*{Acknowledgments}
This project has received funding from the European Research Council (ERC) under the European Union's Horizon 2020 research and innovation programme under grant agreement No 853459. The authors gratefully acknowledge the computer resources at ARTEMISA, funded by the European Union ERDF and Comunitat Valenciana as well as the technical support provided by the Instituto de Física Corpuscular, IFIC (CSIC-UV). This research is supported by a recognition 2021SGR-Cat (01266 LQMC) from AGAUR (Generalitat de Catalunya). We appreciate the discussion with Janis Goldzycher and his suggestions on using a random baseline.
}

\bibliography{anthology,custom}

\appendix

\section{Datasets}
\label{datasets}

\tref{tab:datasets} summarizes the dataset main statistics. 
Unless stated otherwise, the datasets were sourced from HuggingFace Datasets platform \citep{lhoest-etal-2021-datasets}. The primary language of analysis is English, except for Swahili News.

\begin{table}[htbp]
\centering
\begin{tabular}{@{\hspace{0.3em}}l@{\hspace{0.5em}} rrrr@{\hspace{0.3em}}}
\toprule
Dataset & $|\mathcal{Y}|$ & Prior & len. & \# train / test \\
\midrule
IMDB & 2 & 0.5 & 233 & 25k / 25k \\
Wiki Toxic & 2 & 0.096 & 68 & 160k / 64k \\
AG News & 4 & $1/|\mathcal{Y}|$ & 38 & 120k / 7.6k \\
Swahili News & 6 & imb. & 327 & 22k / 7k \\
Arise News & 6 & imb. & 30 & 4.7k / 828 \\
News Articles & 8 & imb. & 835 & 3k / 745 \\
Advers. NLI & 3 & imb. & 63 & 45k / 1k \\
RTE & 2 & 0.496 & 51 & 2.5k / 277 \\
\bottomrule
\end{tabular}
\caption{A summary of the datasets, including number of classes, class distribution, average sequence length, and partition sizes.}
\label{tab:datasets}
\end{table}

\section{Supplementary Results}
\label{extra}

\tref{tab:probing} presents the results regarding the ablation study in \sref{representations}.

The results table concerning the ablation study in \sref{mask_sep} can be found in \tref{tab:mask_sep}.

\begin{table*}[htbp]
\begin{subtable}{\linewidth}
\centering
\begin{tabular}{@{\hspace{0em}}c@{\hspace{.6em}}l@{\hspace{.6em}} lllll@{\hspace{0em}}}
\toprule
\multirow{2.5}{*}{Dataset} & \multirow{2.5}{*}{Repr.} & \multicolumn{5}{c}{Prompt} \\
\cmidrule(r){3-7}
 & & None & Random & Toxicity & Sentiment & Topic \\
\cmidrule(lr){1-7}
\multirow{14}{*}{\begin{tabular}{@{}c@{}}WikiToxic\\(F1$_+$\%)\end{tabular}} & BERT$_1^{\mu}$ & 60.65 & 61.11\pValNR{**0.0}{} & \secondBest{61.55}\pValNR{**0.0}{**0.0015} & \best{61.57}\pValNR{**0.0}{**0.001} & 61.09\pValNR{**0.001}{0.5435} \\
 & BERT$_2^{\mu}$ & 60.16 & 60.56\pValNR{**0.0}{} & \best{61.00}\pValNR{**0.0}{**0.0} & \secondBest{60.97}\pValNR{**0.0}{**0.001} & 60.48\pValNR{**0.006}{0.7135} \\
 & BERT$_1^{\texttt{CLS}}$ & 58.64 & 59.39\pValNR{**0.0}{} & \secondBest{59.97}\pValNR{**0.0}{**0.0} & \best{60.22}\pValNR{**0.0}{**0.0} & 59.94\pValNR{**0.0}{**0.0} \\
 & BERT$_2^{\texttt{CLS}}$ & 59.27 & 59.45\pValNR{0.146}{} & 59.29\pValNR{0.4595}{0.797} & \best{59.95}\pValNR{**0.0}{**0.002} & \secondBest{59.83}\pValNR{**0.001}{*0.014} \\
 & RoBERTa$_1^{\mu}$ & 61.17 & \best{61.87}\pValNR{**0.0}{} & \secondBest{61.68}\pValNR{**0.0}{0.9375} & 61.24\pValNR{0.2725}{!!1.0} & 61.50\pValNR{**0.0045}{!!0.997} \\
 & RoBERTa$_2^{\mu}$ & 61.24 & \best{62.07}\pValNR{**0.0}{} & \secondBest{61.93}\pValNR{**0.0}{0.8455} & 61.51\pValNR{*0.013}{!!1.0} & 61.66\pValNR{**0.0}{!!1.0} \\
 & GPT-2$_1^{\mu}$ & \best{60.25} & 59.28\pValNR{!!1.0}{} & 59.63\pValNR{!!1.0}{**0.0025} & 59.64\pValNR{!!1.0}{**0.0005} & \secondBest{59.84}\pValNR{!!0.997}{**0.0} \\
 & GPT-2$_1^w$ & \best{60.58} & 60.00\pValNR{!!1.0}{} & 59.93\pValNR{!!1.0}{0.7065} & 60.39\pValNR{0.9185}{**0.001} & \secondBest{60.54}\pValNR{0.621}{**0.0} \\

\cmidrule(lr){1-7}
\multirow{9}{*}{\begin{tabular}{@{}c@{}}IMDB\\(acc\%)\end{tabular}} & BERT$_1^{\mu}$ & 86.90 & \secondBest{87.07}\pValNR{**0.007}{} & 87.05\pValNR{*0.0135}{0.579} & \best{87.28}\pValNR{**0.0}{**0.001} & 86.95\pValNR{0.2565}{!0.952} \\
 & BERT$_2^{\mu}$ & 87.14 & 87.43\pValNR{**0.0}{} & 87.40\pValNR{**0.0}{0.6405} & \best{87.52}\pValNR{**0.0}{0.1105} & \secondBest{87.44}\pValNR{**0.0}{0.4545} \\
 & BERT$_1^{\texttt{CLS}}$ & \best{85.70} & 85.58\pValNR{0.9075}{} & 85.51\pValNR{!0.9885}{0.796} & \secondBest{85.64}\pValNR{0.739}{0.191} & 85.53\pValNR{!0.9845}{0.7175} \\
 & BERT$_2^{\texttt{CLS}}$ & 84.35 & 84.42\pValNR{0.214}{} & 84.36\pValNR{0.4385}{0.7115} & \best{84.57}\pValNR{**0.006}{*0.04} & \secondBest{84.48}\pValNR{0.071}{0.252} \\
 & RoBERTa$_1^{\mu}$ & \best{90.73} & 90.51\pValNR{!!1.0}{} & 90.17\pValNR{!!1.0}{!!1.0} & 90.48\pValNR{!!0.9995}{0.652} & \secondBest{90.53}\pValNR{!!0.997}{0.363} \\
 & RoBERTa$_2^{\mu}$ & \best{91.19} & 91.03\pValNR{!!0.9955}{} & 90.88\pValNR{!!1.0}{!!0.9955} & \secondBest{91.05}\pValNR{!!0.996}{0.315} & 90.98\pValNR{!!0.9995}{0.7825} \\
 \bottomrule
\end{tabular}
\end{subtable}
\caption{Probe performance of prompt embeddings. We report best score in bold and second-best in italics, along with statistical significance w.r.t. no prompt (magenta) and w.r.t. random prompts (blue), indicating $p < 0.05$ or $p < 0.01$.}
\end{table*}

\begin{table*}[htbp]\ContinuedFloat
\begin{subtable}{\linewidth}
\centering
\begin{tabular}{@{\hspace{0em}}c@{\hspace{.6em}}l@{\hspace{.6em}} lllll@{\hspace{0em}}}
\toprule
\multirow{2.5}{*}{Dataset} & \multirow{2.5}{*}{Repr.} & \multicolumn{5}{c}{Prompt} \\
\cmidrule(r){3-7}
 & & None & Random & Toxicity & Sentiment & Topic \\
\cmidrule(lr){1-7}
\multirow{9}{*}{\begin{tabular}{@{}c@{}}AG News\\(acc\%)\end{tabular}} & BERT$_1^{\mu}$ & 90.37 & 90.36\pValNR{0.5455}{} & \best{90.46}\pValNR{0.115}{0.1085} & \secondBest{90.43}\pValNR{0.204}{0.1825} & 90.38\pValNR{0.4185}{0.355} \\
 & BERT$_2^{\mu}$ & 90.30 & \best{90.51}\pValNR{**0.007}{} & \secondBest{90.40}\pValNR{0.109}{0.8845} & 90.39\pValNR{0.143}{0.9275} & 90.35\pValNR{0.2895}{!0.965} \\
 & BERT$_1^{\texttt{CLS}}$ & 89.03 & 89.26\pValNR{*0.02}{} & \best{89.65}\pValNR{**0.0}{**0.0005} & \secondBest{89.60}\pValNR{**0.0}{**0.0005} & 89.56\pValNR{**0.0}{**0.0055} \\
 & BERT$_2^{\texttt{CLS}}$ & 87.79 & 87.99\pValNR{0.0645}{} & 88.19\pValNR{**0.0005}{0.057} & \best{88.39}\pValNR{**0.0}{**0.0015} & \secondBest{88.26}\pValNR{**0.0005}{*0.022} \\
 & RoBERTa$_1^{\mu}$ & 91.37 & \secondBest{91.40}\pValNR{0.3375}{} & \best{91.42}\pValNR{0.215}{0.344} & 91.37\pValNR{0.426}{0.5765} & 91.37\pValNR{0.453}{0.589} \\
 & RoBERTa$_2^{\mu}$ & \secondBest{90.51} & \best{90.58}\pValNR{0.218}{} & 90.40\pValNR{0.9325}{!0.9845} & 90.39\pValNR{0.937}{!0.9835} & 90.49\pValNR{0.6135}{0.843} \\

\begin{tabular}{@{}c@{}}News Art.\\(acc\%)\end{tabular} & BERT$_2^{\mu}$ & \best{89.50} & 88.89\pValNR{!0.975}{} & 88.94\pValNR{!0.9655}{0.424} & \secondBest{89.22}\pValNR{0.8225}{0.102} & 88.85\pValNR{!0.9825}{0.561} \\
\begin{tabular}{@{}c@{}}Arise News\\(F1\%)\end{tabular} & BERT$_2^{\mu}$ & \best{82.39} & 81.70\pValNR{0.9115}{} & 81.51\pValNR{!0.968}{0.658} & 81.80\pValNR{0.869}{0.4265} & \secondBest{82.19}\pValNR{0.666}{0.175} \\
\begin{tabular}{@{}c@{}}Swh. News\\(F1\%)\end{tabular} & BERT$_2^{\mu}$ & 67.59 & \best{68.64}\pValNR{**0.0005}{} & 68.50\pValNR{**0.0}{0.727} & 68.37\pValNR{**0.001}{0.8815} & \secondBest{68.62}\pValNR{**0.0}{0.486} \\

\cmidrule(lr){1-7}
 & & None & Random & Toxicity & Sentiment & NLI \\
\cmidrule(r){3-7}
\multirow{6}{*}{\begin{tabular}{@{}c@{}}Adv. NLI\\(F1\%)\end{tabular}} & BERT$_1^{\mu}$ & 32.71 & \secondBest{33.61}\pValNR{0.053}{} & 32.73\pValNR{0.4745}{0.921} & 32.65\pValNR{0.53}{0.95} & \best{33.85}\pValNR{0.0525}{0.3585} \\
 & BERT$_2^{\mu}$ & \secondBest{33.83} & 32.87\pValNR{0.94}{} & 32.89\pValNR{0.9415}{0.49} & 32.85\pValNR{0.932}{0.5175} & \best{34.63}\pValNR{0.1305}{**0.0015} \\
 & RoBERTa$_1^{\mu}$ & 32.27 & \secondBest{32.84}\pValNR{0.1665}{} & 32.53\pValNR{0.314}{0.721} & \secondBest{32.84}\pValNR{0.157}{0.4915} & \best{33.19}\pValNR{0.0885}{0.299} \\
 & RoBERTa$_2^{\mu}$ & 33.65 & 33.47\pValNR{0.632}{} & 33.79\pValNR{0.398}{0.2535} & \secondBest{33.97}\pValNR{0.2715}{0.1695} & \best{34.19}\pValNR{0.2075}{0.1385} \\
\cmidrule(lr){1-1}
\multirow{3}{*}{\begin{tabular}{@{}c@{}}RTE\\(acc\%)\end{tabular}} & BERT$_1^{\mu}$ & \best{51.26} & 48.32\pValNR{!0.9865}{} & 48.65\pValNR{!0.97}{0.3945} & 47.84\pValNR{!!0.994}{0.652} & \secondBest{50.32}\pValNR{0.744}{0.089} \\
 & BERT$_2^{\mu}$ & \best{49.82} & 48.87\pValNR{0.747}{} & 49.30\pValNR{0.6415}{0.365} & 47.44\pValNR{0.95}{0.841} & \secondBest{49.79}\pValNR{0.481}{0.265} \\
\bottomrule
\end{tabular}
\end{subtable}
\caption{(Continued.) Probe performance of prompt embeddings.}
\label{tab:probing}
\end{table*}

\begin{table*}[htbp]
\centering
\begin{tabular}{cl lllll}
\toprule
\multirow{2.5}{*}{Dataset} & \multirow{2.5}{*}{Repr.} & \multicolumn{5}{c}{Prompt} \\
\cmidrule(lr){3-7}
 & & None & Random & Toxicity & Sentiment & Topic \\
\cmidrule(lr){1-7}
\multicolumn{7}{c}{\textit{Masked Prompt}} \\
\multirow{2.5}{*}{\begin{tabular}{@{}c@{}}Wiki Toxic\\(F1$_1$\%)\end{tabular}} & BERT$_1^{\mu}$ & 60.65 & 60.95\pValNR{*0.0145}{} & \best{61.17}\pValNR{**0.0}{*0.0445} & \secondBest{61.16}\pValNR{**0.0005}{*0.0485} & 61.07\pValNR{**0.0}{0.1905} \\
 & BERT$_2^{\mu}$ & 60.16 & 60.65\pValNR{**0.0}{} & \best{60.91}\pValNR{**0.0}{*0.026} & \secondBest{60.75}\pValNR{**0.0}{0.228} & 60.49\pValNR{**0.0055}{0.881} \\
\multirow{2.5}{*}{\begin{tabular}{@{}c@{}}IMDB\\(acc\%)\end{tabular}} & BERT$_1^{\mu}$ & 86.90 & 87.01\pValNR{*0.0475}{} & \secondBest{87.06}\pValNR{*0.01}{0.22} & \best{87.18}\pValNR{**0.0}{**0.003} & 87.03\pValNR{*0.0405}{0.3405} \\
 & BERT$_2^{\mu}$ & 87.14 & \secondBest{87.45}\pValNR{**0.0}{} & 87.28\pValNR{*0.0165}{!!0.996} & 87.42\pValNR{**0.0}{0.676} & \best{87.48}\pValNR{**0.0}{0.3735} \\
\multirow{2.5}{*}{\begin{tabular}{@{}c@{}}AG News\\(acc\%)\end{tabular}} & BERT$_1^{\mu}$ & \secondBest{90.37} & 90.30\pValNR{0.7485}{} & \best{90.40}\pValNR{0.3495}{0.1285} & \best{90.40}\pValNR{0.3575}{0.119} & 90.34\pValNR{0.588}{0.303} \\
 & BERT$_2^{\mu}$ & 90.30 & \best{90.36}\pValNR{0.236}{} & 90.29\pValNR{0.5165}{0.776} & 90.32\pValNR{0.3765}{0.6535} & \secondBest{90.33}\pValNR{0.3465}{0.5975} \\

\cmidrule(lr){1-7}
\multicolumn{7}{c}{\textit{Separator}} \\
\multirow{2.5}{*}{\begin{tabular}{@{}c@{}}Wiki Toxic\\(F1$_1$\%)\end{tabular}} & BERT$_1^{\mu}$ & 60.65 & 61.29\pValNR{**0.0}{} & \best{61.69}\pValNR{**0.0}{**0.0005} & \secondBest{61.57}\pValNR{**0.0}{*0.01} & 61.34\pValNR{**0.0}{0.311} \\
 & BERT$_2^{\mu}$ & 60.16 & 60.85\pValNR{**0.0}{} & \best{61.26}\pValNR{**0.0}{**0.0} & \secondBest{60.95}\pValNR{**0.0}{0.1515} & 60.70\pValNR{**0.0}{0.885} \\
\multirow{2.5}{*}{\begin{tabular}{@{}c@{}}IMDB\\(acc\%)\end{tabular}} & BERT$_1^{\mu}$ & \best{86.90} & 86.82\pValNR{0.8355}{} & 86.87\pValNR{0.666}{0.2255} & 86.82\pValNR{0.8165}{0.4565} & \secondBest{86.89}\pValNR{0.55}{0.1375} \\
 & BERT$_2^{\mu}$ & 87.14 & \secondBest{87.37}\pValNR{**0.0015}{} & \best{87.41}\pValNR{**0.0005}{0.2305} & \secondBest{87.37}\pValNR{**0.0015}{0.4535} & 87.23\pValNR{0.113}{!0.9745} \\
\multirow{2.5}{*}{\begin{tabular}{@{}c@{}}AG News\\(acc\%)\end{tabular}} & BERT$_1^{\mu}$ & 90.37 & \best{90.72}\pValNR{**0.0}{} & \secondBest{90.71}\pValNR{**0.0}{0.481} & 90.70\pValNR{**0.0}{0.5095} & 90.62\pValNR{**0.002}{0.89} \\
 & BERT$_2^{\mu}$ & 90.30 & \secondBest{90.42}\pValNR{0.09}{} & 90.36\pValNR{0.248}{0.74} & 90.41\pValNR{0.083}{0.4625} & \best{90.45}\pValNR{*0.035}{0.303} \\

\cmidrule(lr){1-7}
\multicolumn{7}{c}{\textit{Masked Prompt and Separator}} \\
\multirow{2.5}{*}{\begin{tabular}{@{}c@{}}Wiki Toxic\\(F1$_1$\%)\end{tabular}} & BERT$_1^{\mu}$ & 60.65 & 61.02\pValNR{**0.001}{} & \best{61.44}\pValNR{**0.0}{**0.0} & \secondBest{61.39}\pValNR{**0.0}{**0.0} & 61.01\pValNR{**0.003}{0.5215} \\
 & BERT$_2^{\mu}$ & 60.16 & 60.93\pValNR{**0.0}{} & \best{61.16}\pValNR{**0.0}{*0.016} & \secondBest{61.04}\pValNR{**0.0}{0.1305} & 60.57\pValNR{**0.0}{!!0.9985} \\
\multirow{2.5}{*}{\begin{tabular}{@{}c@{}}IMDB\\(acc\%)\end{tabular}} & BERT$_1^{\mu}$ & 86.90 & 86.86\pValNR{0.6875}{} & \best{86.93}\pValNR{0.371}{0.113} & 86.86\pValNR{0.677}{0.4375} & \secondBest{86.91}\pValNR{0.4325}{0.1765} \\
 & BERT$_2^{\mu}$ & 87.14 & \secondBest{87.32}\pValNR{*0.012}{} & \best{87.40}\pValNR{**0.0}{0.073} & 87.31\pValNR{*0.0145}{0.5185} & 87.18\pValNR{0.311}{!0.98} \\
\multirow{2.5}{*}{\begin{tabular}{@{}c@{}}AG News\\(acc\%)\end{tabular}} & BERT$_1^{\mu}$ & 90.37 & \secondBest{90.73}\pValNR{**0.0}{} & 90.70\pValNR{**0.0}{0.647} & \best{90.74}\pValNR{**0.0}{0.413} & 90.69\pValNR{**0.0}{0.6855} \\
 & BERT$_2^{\mu}$ & 90.30 & \secondBest{90.37}\pValNR{0.206}{} & 90.30\pValNR{0.489}{0.797} & \best{90.39}\pValNR{0.155}{0.3515} & \best{90.39}\pValNR{0.157}{0.358} \\
\bottomrule
\end{tabular}
\caption{Probe performance comparison using representations with masked prompt, using a separator between prompt and sample, or both.}
\label{tab:mask_sep}
\end{table*}

\end{document}